\documentclass[letterpaper, 10 pt, conference]{ieeeconf}  

\IEEEoverridecommandlockouts                              

\usepackage{graphicx} 
\usepackage{amsmath} 
\usepackage{cite}
\usepackage[hidelinks]{hyperref}

\title{\LARGE \bf
Spin Detection in Robotic Table Tennis*
}

\author{Jonas Tebbe$^{1}$, Lukas Klamt$^{1}$, Yapeng Gao$^{1}$, and Andreas Zell$^{1}$
\thanks{*This work was supported in parts by the Vector Stiftung and KUKA}
\thanks{$^{1}$Jonas Tebbe, Lukas Klamt, Yapeng Gao and Andreas Zell are with the Cognitive Systems group, Computer Science Department,
        University of Tuebingen, 72076 Tuebingen, Germany
        {\tt\small [jonas.tebbe, yapeng.gao, andreas.zell]@uni-tuebingen.de, lukas-raphael.klamt@student.uni-tuebingen.de}}%
}
\begin{document}

\maketitle
\thispagestyle{empty}
\pagestyle{empty}

\begin{abstract}
In table tennis, the rotation (spin) of the ball plays a crucial role. A table tennis match will feature a variety of strokes. Each generates different amounts and types of spin. To develop a robot that can compete with a human player, the robot needs to  detect spin, so it can plan an appropriate return stroke. In this paper we compare three methods to estimate spin. The first two approaches use a high-speed camera that captures the ball in flight at a frame rate of 380 Hz. This camera allows the movement of the circular brand logo printed on the ball to be seen. The first approach uses background difference to determine the position of the logo. In a second alternative, we train a CNN to predict the orientation of the logo. The third method evaluates the trajectory of the ball and derives the rotation from the effect of the Magnus force. This method gives the highest accuracy and is used for a demonstration. Our robot successfully copes with different spin types in a real table tennis rally against a human opponent. 

\end{abstract}

\section{INTRODUCTION}

One of the most difficult tasks when playing table tennis is judging the amount of spin on a ball. 
To achieve the goal of beating human players of different levels, a table tennis robot needs to be able to accurately predict spin. A lot of prior knowledge is required to assign the right spin to a shot. The major factor used by human players to judge spin is the opponent's stroke. It is, however, difficult to detect stroke movement with a camera. Such an approach would also require training with a number of different people and rackets. 

Some professional players with excellent eyesight are able to see the rotation of the ball from the movement of the brand logo. By recording the ball with high-speed cameras, it is possible to identify markers on the ball and detect its rotation. This is the most common approach in the literature. Tamaki et al. \cite{Tamaki2012} use black lines on the ball for tracking. Zhang et al. \cite{Zhang2014,Zhang2015} use the logo printed on the ball. 

\begin{figure}[t]
    \centering
    \includegraphics[width=0.45\textwidth]{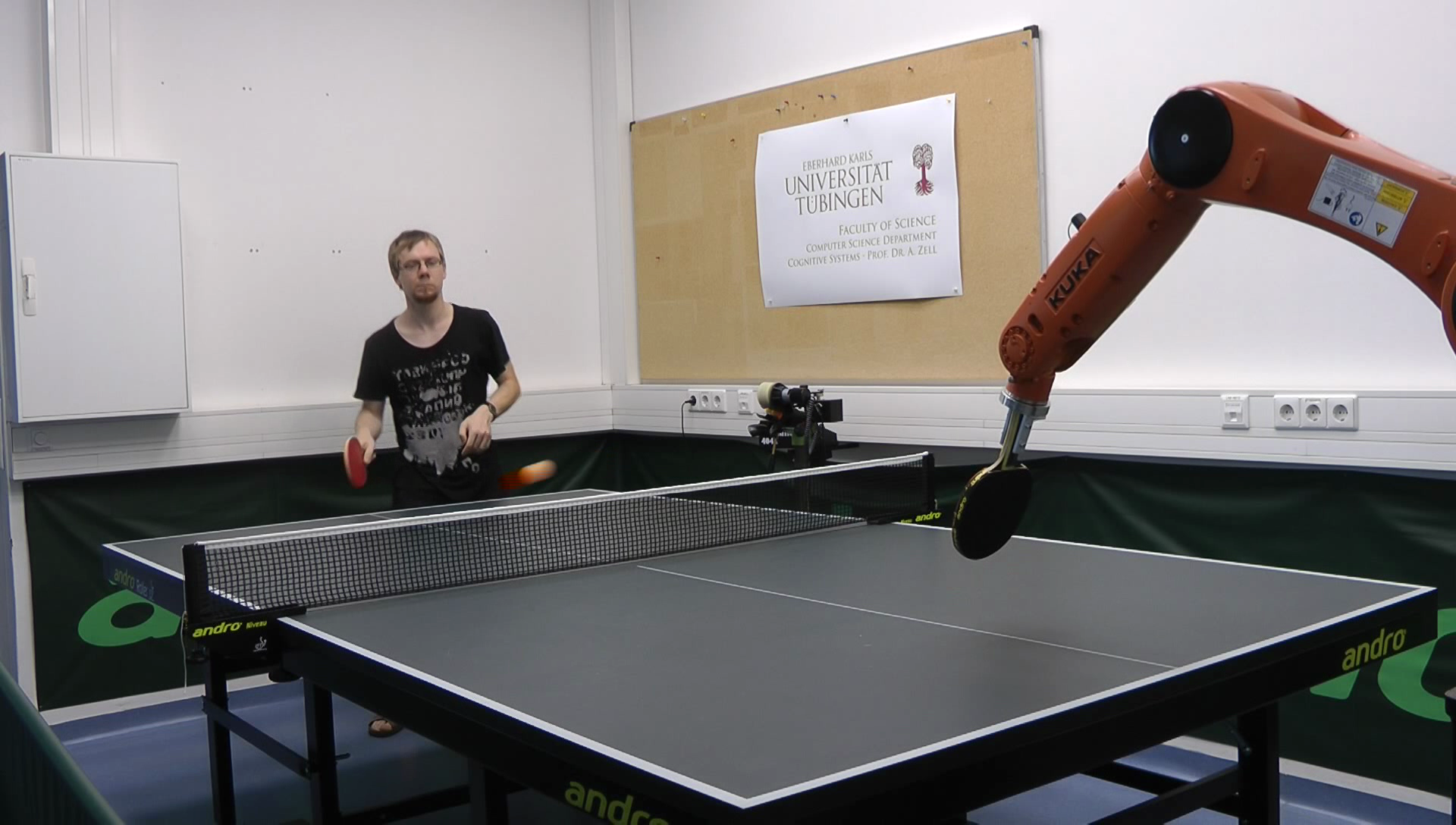}
    \caption{Spin Detection used to return balls with high rotation on a real robot. Supplementary Video: \url{https://youtu.be/SjE1Ptu0bTo}}
    \label{fig:overlays}
\end{figure}

Another promising approach is to use measurements of the ball's trajectory to determine spin. Huang et al. \cite{Huang2011} used a similar approach, involving a physical force model which included the Magnus force, to determine the rotation of the ball. Zhao et al. \cite{Zhao2015,Zhao2017} replace the norm of the velocity necessary to calculate the air resistance. Thus, a differential equation can be solved and one can fit the speed and spin values. Blank et al. \cite{Blank2017} capture stroke motion using an IMU mounted on the bat to predict the rotation of the ball. Gao et al. \cite{yapeng2019} track the table tennis bat using stereo cameras and use a neural network to classify the different types of strokes.

Our goal is to develop a table tennis robot that can successfully return spin strokes from a human opponent. To achieve this we introduce and evaluate three different methods for spin detection using the movement of the ball's logo or its trajectory. 
Our key contributions are summarized by the following:
\begin{itemize}
\item State of the art spin detection using logo extraction is improved by circular segment fitting.
\item A CNN is trained on ball images outperforming standard logo extraction methods. For training and evaluation of the CNN a dataset of 4656 images with manual logo orientation label is created. 
\item Fitting the different forces to the ball trajectory gives a robust spin estimation independent of the ball's logo. 
\item The trajectory fitting is employed on a real robot system and gives convincing results playing against a human opponent, featuring a diversity of strokes (and corresponding spin types) from both human and robot. To our knowledge this level of stroke adaption has not yet been shown by any other robot.
\end{itemize}

Going beyond the topic of this research work, these methods could have an impact on spin detection in other research areas, especially research focusing on sports with fast flying and strongly rotating balls, like tennis, baseball or football. As well as developing robots for these sports, spin detection can also be used for match analysis or for evaluating and improving player technique. There are various general robotic applications where it is necessary to determine the rotation of objects. In the case of table tennis, processing time is the key factor in determining whether or not the application will be successful. In modern highly-dynamic robotic systems, time-optimized object pose detection is essential, e.g. when grasping objects in human-robot collaboration or during autonomous driving in high-speed traffic.

\section{Spin Estimation from the Brand Logo by Background Subtraction}

A PointGrey Grasshopper3 camera is mounted on the ceiling above the center of the tablet tennis table. The camera can achieve 162 fps at full resolution (1920 x 1200). A very high ball spin exceeds 100 revolutions per second. In this case the ball's brand logo would be visible only every second frame. We therefore selected a ROI of 1920 x 400 and record at 380 fps, which is possible with this camera type.  The exposure time was $0.25$ ms.

\subsection{Ball Detection}
\begin{figure}[thpb]
	\centering
	\includegraphics[width=0.4\textwidth]{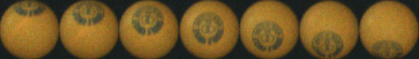}\\
	\vspace{0.1cm}
	\includegraphics[width=0.4\textwidth]{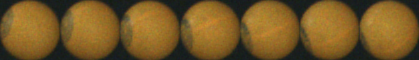}
    \caption{Top row: A sequence of ball images in which the rotation of the brand logo is fully visible. Bottom row: The logo is also visible throughout the sequence, but the movement at the edge of the ball's image is more difficult to see.}
	\label{fig:logo-sequence}
\end{figure}

The ball is extracted from the image using a frame difference method taken from \cite{Tebbe2018}. 
Figure \ref{fig:logo-sequence} presents two example sequences of cropped ball images showing the movement of the brand logo.

\subsection{Logo Contour Detection}
\begin{figure}[thpb]
\centering
\def\bewidth{1.05cm}
\begin{tabular}{cccccc}
\includegraphics[width=\bewidth]{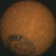}&
\includegraphics[width=\bewidth]{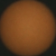}&
\includegraphics[width=\bewidth]{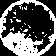}& 
\includegraphics[width=\bewidth]{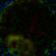}&
\includegraphics[width=\bewidth]{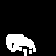}&
\includegraphics[width=\bewidth]{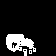}\\
(a)&(b)&(c)&(d)&(e)&(f)
\end{tabular}
\caption{logo detection process using motion and color features: (a) current frame, (b) ball without logo, (c) color threshold of a, (d) difference of a and b (e) binary threshold of d, (f) bitwise or of (c) and (e)}
\label{fig:brand-extraction}
\end{figure}

Ball detection yields an image containing only the ball at a size of around 70 x 70 pixels. The process of marking all the pixels that belong to the brand logo is described in figure \ref{fig:brand-extraction}.
For all pixels of the logo contour, we want to know the 3D positions on the ball.

\subsection{3D Projection}
The ball extraction also gives the radius of the ball in pixels. This is calculated by fitting a circle to the ball blob. For each contour pixel, we first calculate its position relative to the ball's center. The $x$ and $y$ components are then divided by the ball's radius. Since our 3D point must lie on the unit sphere, the $z$ component can be derived from 
\[
1 = x^2 + y^2 + z^2.
\]

\subsection{Brand Logo Center}

The next step involves calculating the logo center. In our first approach, we simply normalize the average of all 3D contour points. This does not take into account the fact that contour points closer to the ball's center in the image are more frequent. The centroid can also be calculated iteratively using Ritter's bounding sphere \cite{Ritter1990} on the 3D contour points. Normalization projects the centroid onto the unit sphere. As this did not significantly boost accuracy, we used the first approach, which was faster. 

\begin{figure}[h]
\centering
\includegraphics[width=0.45\linewidth]{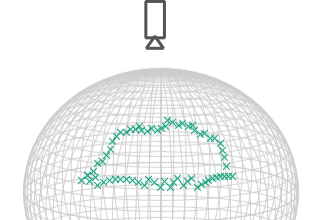}
\includegraphics[width=0.45\linewidth]{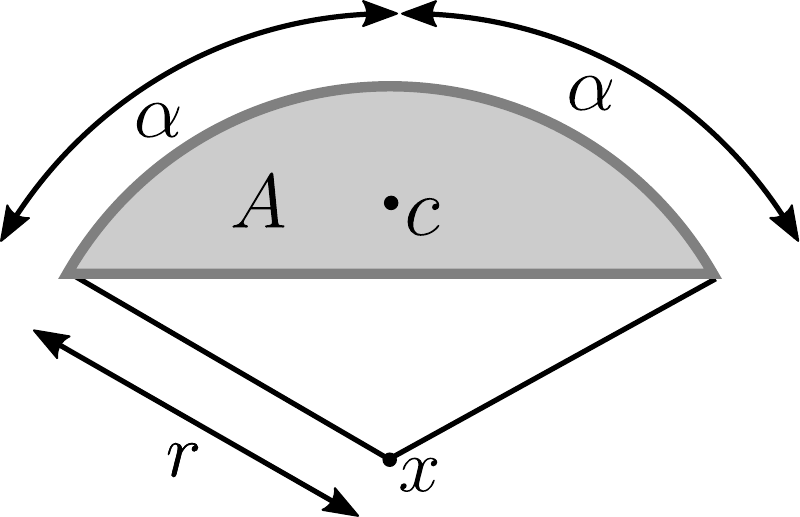}
\caption{Left: 3d projected logo contour for a partially visible ball seen from the top. Right: Schematic representation of a circular segment}
\label{fig:circluar-segment}
\end{figure}

On the camera images only one side of the ball is visible. Therefore, brand logos may be only partially in view when they are located at the edge of the shown area. The left of figure \ref{fig:circluar-segment} shows a contour transformed into the 3D ball coordinate space for such an edge case. In this case the contour does not form a circle but a 2D circle segment, so the center position cannot be obtained as before. 

We approximate the area $A$ as $\pi$ times the average distance from the contour (green crosses in figure \ref{fig:circluar-segment} left) to the centroid.
We know the actual radius $r$ of the logo from measurements. We can therefore derive the distance from the centroid $c$ to the real center $x$ from the area $A$ \cite{wiki_centroids}: \begin{align*}
\label{eq:segment-area}
A &= \frac{r^2}{2} ( 2 \alpha - \sin(2\alpha)) \\
d(x,c) &= \frac{4 r \sin^3(\alpha)}{3(2\alpha-sin(2\alpha))}.
\end{align*}
To get the real 3D center we rotate the centroid $c$ by angle $\beta = 360^{\circ} d(x,c) / 2\pi r$ around the axis $(0,0,1)$. The circular segment fitting stabilizes the spin detection for the challenging edge case compared to the original approach of Zhang et al. \cite{Zhang2014}.

\subsection{Fitting Rotation}
\label{fitting_rotation}
\begin{figure}[htpb]
	\centering
	\includegraphics[width=0.4\textwidth]{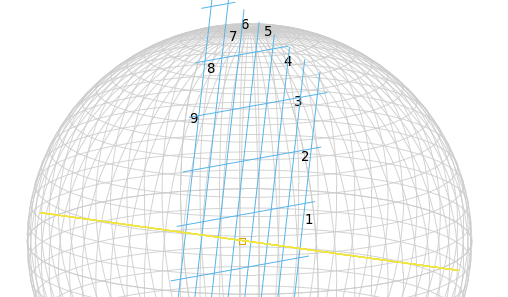}
	\caption{Detected ball positions displayed in chronological order. The fitted plane is visualized by a grid of lines.}
	\label{fig:plane-fitting}
\end{figure}

After processing $10$ to $30$ images captured every $1/380 s$ from the ball's trajectory as described above, we can estimate the ball's spin. For this purpose, we fit a plane through the center points.  The fitted plane should minimize the distance to the points. 
Additionally, the distance of the points to the circle created by intersecting the plane with the ball, represented as the unit sphere, should be minimized.
An example is shown in figure \ref{fig:plane-fitting}.

To get the angular velocity, we project the logo positions onto the plane and calculate the angle $\omega$ between two consecutive logo positions. If the logo was not found on two or more successive images, the ball has made a half revolution. The rotation is therefore described not by the short angle $\omega$ between the points before and after but by the large angle $360^\circ - \omega$. At the end we have a sequence of the accumulated angles and fit a regression line to the sequence. The gradient of that line gives us the angular velocity.

\section{Spin Estimation by CNN on ball image}

Our second approach uses a Convolutional Neural Network (CNN) to estimate the visibility of the logo and the 3D pose of the ball. We then use the same algorithm as in section \ref{fitting_rotation} to estimate the rotation axis and angular velocity. 
\subsection{Dataset}
\label{sec:dataset}
To train and test the network, a total of 4656 images were recorded using our PointGrey Grasshopper3 camera. The images were cropped around the table tennis ball to have a fixed size of 60 x 60 pixels. 46.7\% were labeled as having no visible brand logo. The ball's pose was labeled with the help of a 3D scene containing a ball with realistic logo texture. The 3D scene was modeled with the open-source 3D computer graphics software Blender \cite{Blender}. Each real ball image was placed transparently over the scene. Next, the 3D ball model can be optimized to fit the actual image and the pose can be read out by the Blender Python API.

\begin{table*}[ht]
    \centering
    \begin{tabular}{c|c|c|c|c|c|l|l}
          GAP & FC & dropout & train. loss & test loss & classification & geodesic & vector angle  \\
          &  & & cond.-$\mathcal{L}_1$ & cond.-$\mathcal{L}_1$ & accuracy & in deg. & in deg.  \\
          \hline
          1 & - & -   & \textbf{0.0309} & 0.3028 & 0.974 & 33.97 & 7.72 \\
          \hline
- & 3 & -   & 0.0342 & 0.3212 & \textbf{0.975} & 36.32 & 8.01\\
- & 3 & 0.5 & 0.2674 & 0.4288 & 0.974 & 49.86 & 19.81 \\
- & 3 & 0.8 & 0.1347 & 0.3162 & 0.974 & 34.55 & 13.74 \\
\hline
1 & 3 & -   & 0.1006 & 0.2199 & \textbf{0.975} & 24.01 & 5.08 \\
1 & 3 & 0.5 & 0.0976 & 0.2215 & 0.974 & 23.90 & 5.16 \\
1 & 3 & 0.8 & 0.1022 & \textbf{0.2161} & 0.974 & \textbf{23.06} & \textbf{4.89}
    \end{tabular}
    \caption{All models are ResNet architectures. The classification column shows the accuracy of the binary classification task. Geodesic describes the geodesic distance between ground truth label and prediction. Vector angle denotes the metric from Chapter \ref{sec:losses}. The best results are marked in \textbf{bold}.}
    \label{tab:architecture}
\end{table*}

\subsection{Network Architecture}
\label{sec:architecture}
Related work on pose detection with neural networks \cite{Mahendran20173DNetworks, SadeghReal-timeRegistration} favours the residual network architecture from He et al. \cite{HeDeepRecognition}. To use the information for our robot, the model has to run approximately in real-time. Therefore we use the smallest of the ResNet architectures from \cite{HeDeepRecognition} having 18 layers.

Expanding the idea of Mahendran et al. \cite{Mahendran20173DNetworks} we tested networks with two additional fully-connected layers (FC) with 512 neurons each right before the final regressor. This modification should improve the transformation from feature space to pose space. The effectiveness of the additional layers at various dropout rates can be observed in table \ref{tab:architecture}.

\subsection{Network Output}
There are several mathematical representations of a rotation. One can use rotation matrices, Euler angles, axis angles representation, or quaternions. Matrices do not fit as output of our network, as more parameters need to be estimated and one needs to ensure that the result is within the matrix subgroup $SO(3)$ of rotation matrices. With Euler angles it is difficult to represent continuous rotations. As a result, we trained networks to predict the pose of the table tennis ball in either axis angle representation or in quaternions. For either representation, the output is concatenated with a real number for the visibility of the brand logo. The range of the visibility value is $[-1,1]$ to match the $z$-positions away from the camera. In the dataset non-visible logos are labeled as $-1$.

\subsection{Loss Functions}
\label{sec:losses}
The proposed neural network has to learn two tasks simultaneously. It needs to classify whether the brand logo of the ball is visible and predict the pose of the ball. If the logo is not visible, the pose cannot be determined. In this case, the network should not learn any incorrect poses. Hence, we define a conditional loss function that splits the loss into the two tasks:
\[
\mathcal{L} = \mathcal{L}_{classification} + t_{v} ~ \mathcal{L}_{orientation}
\]
where $t_{v}$ denotes the binary ground truth visibility value. For a visible logo, the value is $1$. Otherwise it is $0$. Therefore, we call it conditional loss.

When outputting in axis angle or quaternion representation, we adjust the pose losses for ambiguity. In both representations, the negative value gives the same orientation since the rotation in the opposite direction about the negative axis corresponds to the original rotation. We tested both $L_1$ and $L_2$ norms to get the following conditional losses:
\begin{align*}
\mathcal{L}_1 &= (o_{v} - t_{v})^2 + t_{v} \min\left(\sum_{i=0}^{n} (o_i - t_i)^2, \sum_{i=0}^{n} (o_i + t_i)^2 \right) \\
\mathcal{L}_2 &= |o_{v} - t_{v}| + t_{v} \min\left((\sum_{i=0}^{n} |o_i - t_i|, \sum_{i=0}^{n} |o_i + t_i| \right)
\end{align*}
where $o = (o_1, \cdots, o_{n},o_{v})$ is the output vector of the network and $t = (t_1, \cdots, t_{n},t_{v})$ is the target vector.

A more complex, but fairly exact measurement of the accuracy of rotations is the geodesic distance in $SO(3)$. 
For two rotations this metric returns the angle (from axis-angle representation) of the rotation aligning them both. If $R_1,R_2$ are rotation matrices the geodesic distance is calculated as
$$ d_{GD}(R_1,R_2) = arccos\left(\frac{tr\left(R_1^T R_2 \right) -1}{2} \right)$$
For quaternions $q_1,q_2$ the geodesic distance is computed by
$$d_{GD}(q_1,q_2) = 2 arccos(|<q_1,q_2>|)$$
where $|\cdot|$ denotes the absolute value and $<\cdot,\cdot>$ is the inner product of 4-dimensional quaternion vectors. As before, we define a new loss function
$$ \mathcal{L}_{GD} = |o_v - t_v| + t_{v} ~ d_GD\left(o, t\right).$$

The best performance was achieved by training quaternions with the conditional $\mathcal{L}_2$ loss (see table \ref{tab:expRot}). 

\begin{table}[h]
    \centering
    \begin{tabular}{c|c|c|c}
        loss & rotation& geodesic & vector angle  \\
        function & representation& in deg. & in deg.  \\
        \hline
        cond.-$\mathcal{L}_1$ & quaternions& \textbf{23.06} & \textbf{4.89} \\
        & axis angle  & 27.40 & 9.93\\
        \hline
        & quaternions  & 27.92 & 6.69\\
        cond.-$\mathcal{L}_2$ & axis angle  & 30.90 & 11.20\\
        \hline
        & quaternions  & 43.38 & 13.54\\
        L2 & axis angle  & 37.63 & 17.90\\
        \hline
        Geodesic & quaternions  & 23.49 & 5.97\\
    \end{tabular}
    \caption{Network metrics evaluated on networks trained with different loss functions and rotation representations.}
    \label{tab:expRot}
\end{table}

\subsection{Metrics}

The most difficult part of the rotation for the networks to determine is the logo's orientation about its center. We therefore also want to evaluate the accuracy of the network's prediction of the position of the logo on the ball only, i.e. without considering whether it is rotated in itself. For that we need an additional metric not affected by the orientation. We convert the rotation of the ball to logo positions, represented by points on the unit sphere, by rotating the base logo position $(0,0,1)$. The metric is then the vector angle describing the angle between two positions. 

The network is used on several images of the ball trajectory. For the final spin estimation the poses outputted from the networks are converted to logo positions as described in the previous paragraph. We then use the same algorithm as in section \ref{fitting_rotation} to estimate rotation axis and angular velocity.

\subsection{Training Setup}
The dataset from section \ref{sec:dataset} was split into training and test set with a $4:1$ ratio. As a result, 3725 images were used for training and 931 for testing. The networks were trained with Tensorflow using an Nvidia GeForce GTX 1080 Ti graphics card.

\subsection{Inference Time}
\label{sec:inference-time}
In our scenario it is not just accuracy that matters - time for the evaluation (inference time) is also important. From the camera we get images at $380$ Hz. This gives us a processing time of $2.6$ ms per image for real-time performance. Segmenting the ball out and cropping takes $0.5$ ms, leaving $2.1$ ms for the network. The best network has an inference time of $3.7$ ms on a GTX1080 Ti graphics card. We solve the problem by processing in batches of $5$ images, taking only $5.5$ ms in total due to reduced overhead.


Our best performing network is an 18-layer ResNet plus global average pooling and two fully connected layers trained to output quaternions with conditional $\mathcal{L}_2$ loss. Augmenting the data with $90^\circ$ rotations and Gaussian noise with standard deviation of $5\%$ achieves the following result:

\begin{center}
    \begin{tabular}{ccccc}
          class. acc. & geodesic & vector angle  \\
          0.96 & $20.14^\circ$ & $4.23^\circ$ \\
    \end{tabular}
\end{center}


\section{Spin Estimation from the Trajectory}

In this section we introduce a way of estimating the spin from the trajectory of the ball. We utilize the fact that the rotation of the ball acts on the ball via the Magnus force. Similar work on the topic has been done by Huang et al. \cite{Huang2011}.

\begin{figure}[thpb]
    \centering
    \includegraphics[width=0.4\textwidth]{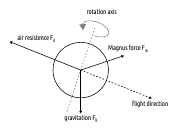}
    \caption{The graphic visualizes the three forces acting on the ball: gravitation pointing downwards, air resistance or drag force in the opposite direction to the flight, and Magnus force perpendicular to the spin axis and flight direction. }
    \label{fig:Forces}
\end{figure}

The forces are depicted in figure \ref{fig:Forces}. The gravitational force $F_g$ is directed towards the ground. The drag force coming from the air resistance acts in the opposite direction to the flight of the ball. The Magnus force is perpendicular to the rotation axis and the flight direction. The acceleration of the ball is therefore calculated by
\begin{align}
\label{eq:accelerations}
\dot{v} = - k_D  \left \| v \right \| v + k_M ~ \omega \times v - 
\begin{pmatrix}
0\\ 
0\\ 
g
\end{pmatrix}.
\end{align}
The notation is shortened with $k_D = - \frac{1}{2} C_D \rho_a A / m$ and $k_M = \frac{1}{2} C_M \rho_a A r / m$, where the constants are the mass of the ball $m=2.7 \text{g}$, the gravitational constant $g = 9.81 \text{m}/\text{s}^2$, the drag coefficient $C_D = 0.4$, the density of the air $\rho_a = 1.29 \text{kg}/\text{m}^3$, the lift coefficient $C_M = 0.6$, the ball radius $r = 0.02 \text{m}$, and the ball's cross-section $A = r^2 \pi$.
For a ball with medium to heavy spin the forces all have similar magnitudes, as can be seen in figure \ref{fig:forces-lookahead}.

\subsection{Fitting}

\begin{figure}[thpb]
	\centering
	\includegraphics[width=0.45\textwidth]{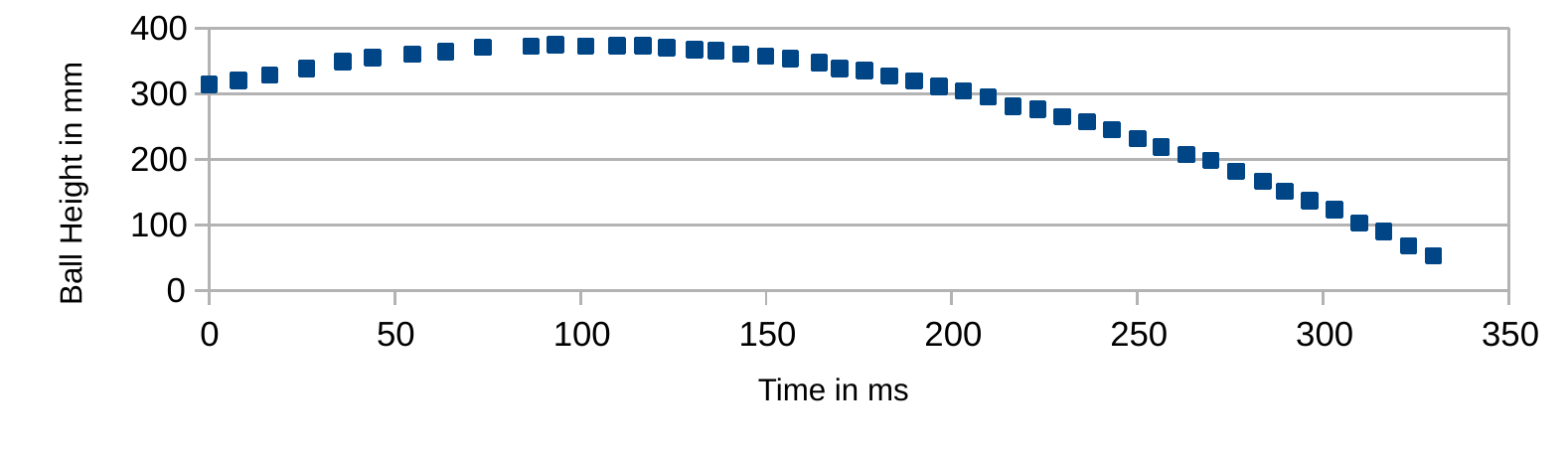}\\
	\includegraphics[width=0.45\textwidth]{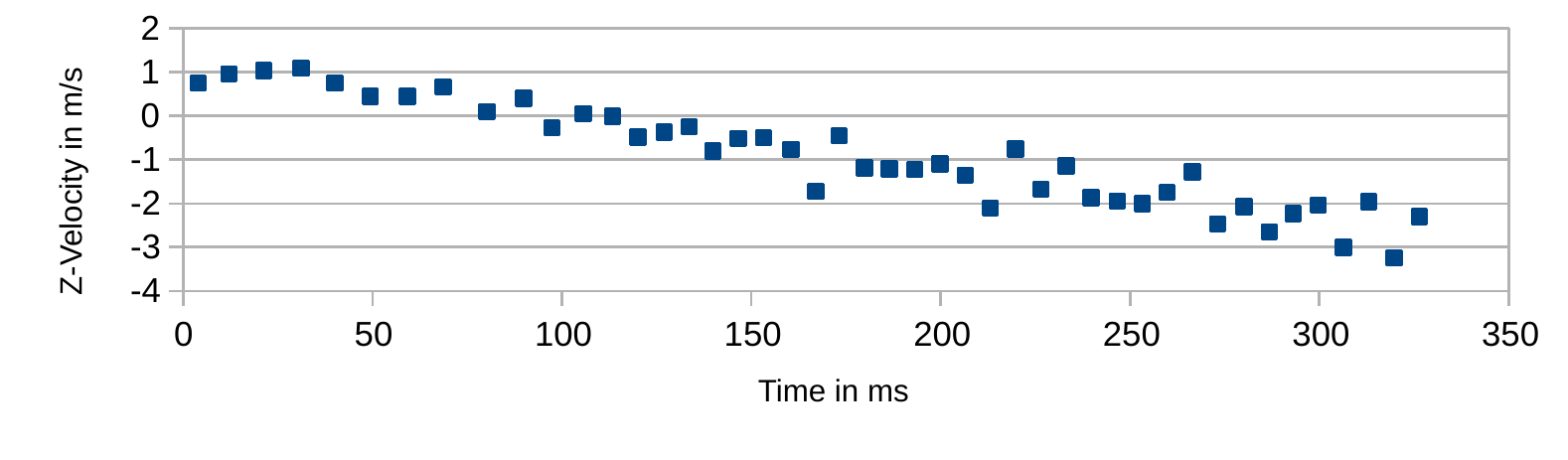}
    \caption{The top diagram shows the height or z-positions for an example trajectory. For the same trajectory the z-velocity, calculated between each pair of neighbouring points, is shown below.}
	\label{fig:two-point-velocity}
\end{figure}

\begin{figure}[thpb]
	\centering
	\includegraphics[width=0.45\textwidth]{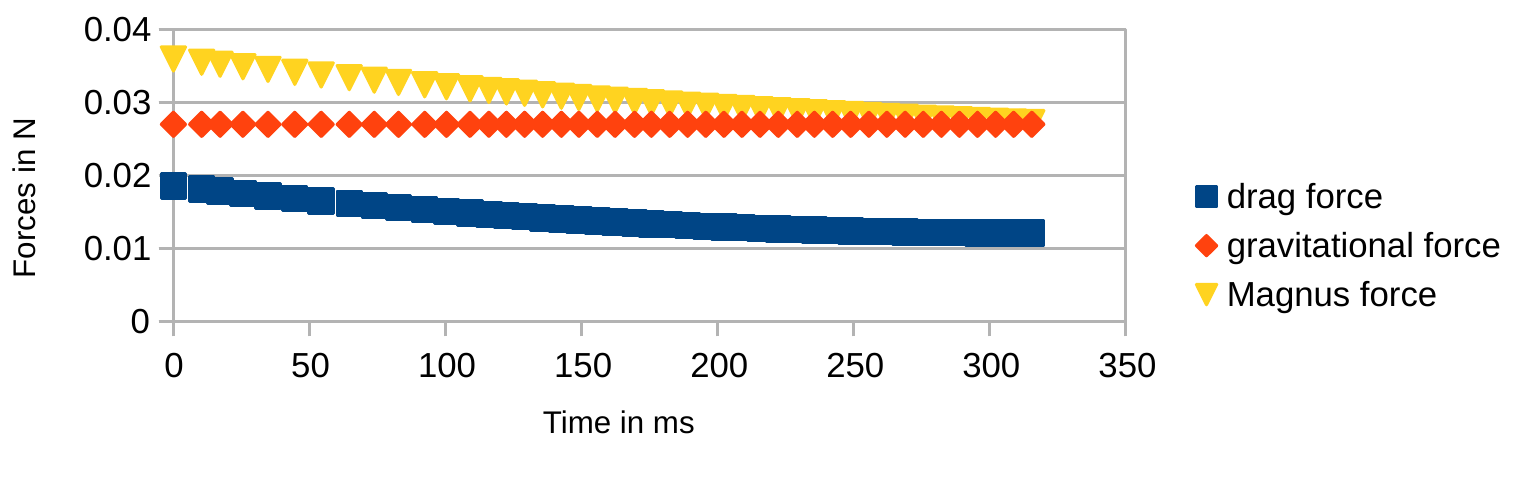}
    \caption{In the diagram the three involved forces for the Magnus fitting are displayed for an example trajectory of a topspin ball.}
	\label{fig:forces-lookahead}
\end{figure}

Given 3D observations $b_1, ... , b_n$ of the ball with ball positions $b_i = (x_i,y_i,z_i)$ at times $t_1, ... , t_n$ we need to estimate velocity and acceleration of the ball to derive the Magnus force. Calculating the velocity between two points is error prone as seen in figure \ref{fig:two-point-velocity}. The problem is solved by fitting a third degree polynomial $P(t) = (P_x(t),P_y(t),P_z(t))$ for each axis using a standard least-squares algorithm. At time step $t$ the velocity is approximated by $P'(t)$ and the total acceleration by $P''(t)$. Rewriting equation \eqref{eq:accelerations} with these approximations yields
\begin{align*}
k_M ~ \omega \times P'(t) = P''(t) + k_D  \left \| P'(t) \right \| P'(t) +   
\begin{pmatrix}
0\\ 
0\\ 
g
\end{pmatrix}.
\end{align*}
Here we assume that the rotation vector $\omega$ is constant within the time of flight. We get this equation for each $t = t_1,...,t_n$. All the equation can be transformed into the equation system $M\omega = a$ with a $m \times 3$ matrix $M$ and an m-dimensional vector of accelerations $a$. We then get a least-squares solution for $\omega$. Note that the acceleration caused by drag force is perpendicular to the Magnus acceleration on the left. As an effect our fitting of $\omega$ does not depend on the coefficient $k_D$ but only on $k_M$ in contrast to other work \cite{Huang2011,Zhao2017}. In figure \ref{fig:forces-lookahead} the forces in each step are displayed for an example trajectory.

\subsection{Preprocessing: Outlier filtering}
The process is error prone to outliers. Even for a slight impact for the fitted trajectory these outliers can produce unrealistic fitted spin values. Especially at the beginning of the trajectory incorrect recognitions can occur when a part of the human body, e.g. the hand, is detected instead of the ball due to its roughly circular shape. For the first 20 balls we select the last 5 balls and make a polynomial fit as above. If the error for the ball $\#14$ is below a specific threshold we start again with balls $\#14$ to $\#19$ otherwise we remove ball $\#14$ as outlier. Repeating this process we remove detected objects which do not belong to the trajectory at the beginning.

With the position $b=P(t_n)$, speed $v=P'(t_n)$ and spin $\omega$ we predict the future trajectory. The improvement for the prediction can be seen in table \ref{tab:bounce-prediction}. We tested backspin, side spin and topspin at three different speed settings with our TTmatic ball throwing machine. For comparison a Kalman filter is used to only predict position and speed without considering the angular velocity of the ball. The statistic includes 90 trajectories in total divided into 10 trajectories each. 
The estimated spin values significantly improves bounce estimation. In contrast to the first two approach using the ball's logo, the spin can used for predicting the future trajectory without adjusting the Magnus coefficient $C_M$. As we divide by it for spin estimation we multiply again for prediction. For the other methods, we found no constant $C_M$ independent of the spin type, which gave good results.
\begin{table}[ht]
\centering
\begin{tabular}{ll|rr|rr}
         &        & \multicolumn{2}{c}{With fitted spin}         & \multicolumn{2}{c}{Without spin value}        \\
         in mm&        & Error      & Stddev & Error        & Stddev \\
\hline
\hline
Backspin & Low   & 10.28            & 5.09   & 36.78              & 6.57   \\
         & Medium & 27.02            & 11.22  & 125.76             & 17.08  \\
         & High   & 43.37            & 32.14  & 170.75             & 25.15  \\
\hline
Sidespin & Low   & 9.68             & 5.56   & 43.15              & 7.99   \\
         & Medium & 16.35            & 10.47  & 82.74              & 13.82  \\
         & High   & 27.99            & 9.80   & 108.24             & 11.23  \\
\hline
Topspin  & Low   & 19.01            & 5.62   & 90.10              & 16.96  \\
         & Medium & 23.36            & 11.24  & 167.17             & 14.76  \\
         & High   & 86.84            & 52.70  & 338.28             & 31.00 
\end{tabular}
\caption{Results on bounce point prediction for balls served from a ball throwing machine with different settings. For each setting we recorded 10 trajectories.}
\label{tab:bounce-prediction}
\end{table}

\section{Comparison}

In this paper, we looked at three algorithms to detect the spin of a table tennis ball. The first two can be compared by evaluating the angular error between the actual and the predicted logo position. The original background subtraction method gives an angular error of $5.77^\circ$. We improved it to $5.06^\circ$ by using circular segment fitting. Our best convolutional neural network reached an error of only $4.23^\circ$. However, both background subtraction methods are faster, with an average processing time of $0.3$ ms, compared to the network inference at $3.7$ ms per image. Batch processing accelerates inference slightly (section \ref{sec:inference-time}).  

For the final spin prediction there are no ground truth values available. Therefore we evaluate how good the algorithms are for the classification of spin types. Using a TTmatic ball throwing machine we recorded 50 trajectories each for 3 spin types and 3 different powers, 9 settings in total. Unfortunately the machine does not allow the speed and rotation of balls to be set independently. Faster spin is therefore accompanied by a higher velocity. The median spin is calculated for each algorithm and setting. This 3D vector defines a cluster in three-dimensional space. Each spin value is then assigned to the nearest cluster center. The accuracy values in table \ref{tab:classfication-result} show how many of the trajectories were correctly classified for each setting.

\begin{table}[b]
\centering
\begin{tabular}{l|r|r|r|r}
Spin type     & Background  & Bg. sub. + & CNN & Trajectory  \\
        &  subtraction & segment fit& &   fitting \\
          \hline
          \hline
Backspin &&&&\\
~ Low   & 88.0\% & 94.0\%      & 96.0\%      & 100.0\%     \\
~ Medium & 84.0\% & 92.0\%      & 94.0\%      & 58.0\%     \\
~ High   & 70.0\% & 86.0\%      & 80.0\%      & 60.0\%     \\
          \hline
Sidespin &&&&\\
~ Low   & 94.0\% & 98.0\%      & 98.0\%      & 100.0\%    \\
~ Medium & 68.0\% & 58.0\%      & 74.0\%      & 94.0\%     \\
~ High   & 60.0\% & 68.0\%      & 66.0\%      & 100.0\%    \\
          \hline
Topspin  &&&&\\
~ Low   & 84.0\% & 90.0\%      & 88.0\%      & 86.0\%     \\
~ Medium & 78.0\% & 86.0\%      & 88.0\%      & 96.0\%     \\
~ High   & 90.0\% & 96.0\%      & 96.0\%      & 100.0\%    \\
          \hline
          \hline
in total  & 79.6\% & 85.3\%      & 86.7\%      & 88.2\%   
\end{tabular}
\caption{Clustering accuracy of all the algorithms.}
\label{tab:classfication-result}
\vspace{-0.5cm}
\end{table}

Surprisingly, the algorithms are similar in accuracy, slightly in favor of the trajectory fitting. A drop in performance is noticeable for balls with a lot of side spin. For this spin type the logo often rotates around itself at the top and hardly changes position. Then the first two variants reached their limit. For the same case appearing on the invisible side the logo cannot be seen and evaluated with these methods. The third algorithm does not suffer from brand logo dependence. However, it had difficulty distinguishing between the medium and high backspin. For these, the trajectories were not different enough leading to two median values relatively close to each other. All in all, a good classification can be achieved with all methods. An improvement would probably achieved by combining an approach using the brand logo with the Magnus force fitting.

\section{Evaluation on a table tennis robot}
The success of spin detection is demonstrated on a KUKA Agilus KR6 R900 robot arm. The table tennis robot system has to respond to different spin types generated by a human opponent. For this demonstration we decided to go with the trajectory Magnus force fitting. It is more accurate, easier to set up and uses fewer resources. No additional camera hardware is necessary since the ball's positions are already captured to predict the trajectory.

We originally developed a table tennis robot system to play without spin \cite{Tebbe2018}. In short, the ball position is extracted from a multi-camera system using color and movement information. Then the ball's 3D position is triangulated. To track the position and velocity of the ball we use an extended Kalman filter. The future trajectory is predicted from this using the force equation \eqref{eq:accelerations}. As soon as we roughly know, where to hit the ball, we move the robot to this position with predefined bat orientation and velocity using a custom driver software and the KR-C4 controller.

In the new spin scenario the return strokes use a different bat orientation. It is given in Euler angles in the order $zyx$. The $z$- and $x$-angle are still only dependent on the $y$-position of the hitting point. However, the $y$-angle $\beta$ is linearly dependent on the $y$-component $\beta_{spin}$ of the fitted rotational velocity of the ball. For a topspin or backspin ball with $\beta_{spin} = \pm 360^\circ /s$, we use a $\beta$ of $28^\circ$ and $-40^\circ$, respectively. For other values of $\beta_{spin}$ we interpolate linearly. At the time of hitting the table tennis racket has a velocity of approximately $1$ m/s in the direction of the human opponent. 

A video demonstration of the experiment is available online\footnote{\url{https://youtu.be/SjE1Ptu0bTo}}. The rubber of the bat is a professional table tennis rubber with high friction. A lot of spin therefore acts on the ball after contact with the bat and our robot is still able to return the ball consistently. As far as we are aware, no other table tennis robot has yet achieved the feat of returning the ball under such challenging conditions. 

\section{Conclusion and future work}
Three methods for spin detection of a table tennis ball were introduced or further enhanced. These approaches were compared. Yielding the best accuracy, the trajectory fitting method is used to generate consistent returning strokes with our KUKA Agilus robot in cooperative, but highly challenging spin-play against a human opponent.

In future, we plan to go from cooperative to competitive strokes. Although our robot control approach is effective for cooperative spin play, it clearly has limits in terms of adaptability. Only the angle of the bat while hitting is adapted using a basic linear model. In a next step, the predicted spin may help to train the speed and orientation of the bat in a more sophisticated way using reinforcement learning.
\addtolength{\textheight}{-12cm}   

\bibliographystyle{IEEEtran}
\bibliography{root}

\end{document}